\documentclass[times,twocolumn,final]{elsarticle}

\usepackage{prletters}
\usepackage{framed,multirow}
\usepackage{amssymb}
\usepackage{latexsym}
\usepackage{url}
\usepackage{amsmath}
\usepackage{booktabs} 
\usepackage{algorithm}
\usepackage{subfigure}
\usepackage[noend]{algpseudocode}
\usepackage{hyperref}

\journal{Pattern Recognition Letters}

\begin{document}

\newcommand{\etal}{\textit{et al}.}
\newcommand{\ie}{\textit{i}.\textit{e}.}
\newcommand{\eg}{\textit{e}.\textit{g}.}

\ifpreprint
  \setcounter{page}{1}
\else
  \setcounter{page}{1}
\fi

\begin{frontmatter}

\title{Compositional Coding Capsule Network with K-Means Routing for Text Classification}

\author[1]{Hao \snm{Ren}} 
\ead{hren17@fudan.edu.cn}
\author[1]{Hong \snm{Lu}\corref{cor1}}
\cortext[cor1]{Corresponding author: 
  Tel.: 86-21-51355528;  
  fax: 86-21-51355558;}
\ead{honglu@fudan.edu.cn}

\address[1]{School of Computer Science, Fudan University, 220 Handan Rd., Yangpu District, Shanghai, 200433, China}

\received{25 March 2022}
\finalform{25 March 2022}
\accepted{25 March 2022}
\availableonline{25 March 2022}
\communicated{Hong Lu}

\begin{abstract}
Text classification is a challenging problem which aims to identify the category of texts. In the process of training, word embeddings occupy a large part of parameters. Under the limitation of limited computing resources, it indirectly limits the ability of subsequent network designs. In order to reduce the number of parameters, the compositional coding mechanism has been proposed recently. Based on this, this paper further explores compositional coding and proposes a compositional weighted coding method. And we apply capsule network to model the relationship between word embeddings, a new routing algorithm, which is based on k-means clustering theory, is proposed to fully mine the relationship between word embeddings. Combined with our compositional weighted coding method and the routing algorithm, we design a neural network for text classification. Experiments conducted on eight challenging text classification datasets show that the proposed method achieves competitive accuracy compared to the state-of-the-art approach with significantly fewer parameters.
\end{abstract}

\begin{keyword}
\MSC 41A05\sep 41A10\sep 65D05\sep 65D17
\KWD Text Classification\sep Capsule Network\sep Compositional Code\sep Deep Learning
\end{keyword}

\end{frontmatter}

\section{Introduction}

Recurrent Neural Networks (RNNs,~\cite{mikolov2010recurrent, elman1990finding, rumelhart1986learning}), including Long Short Term Memory networks (LSTMs,~\cite{hochreiter1997long, gers2000learning}) and Gated Recurrent Units (GRUs,~\cite{chung2014empirical, zhou2016minimal}) have been increasingly applied to many problems in Natural Language Processing (NLP). Text classification is one of the most basic and important tasks in this field.

However, NLP models often require a massive number of parameters for word embeddings, resulting in a large storage or memory footprint. To reduce the number of parameters used in word embeddings without hurting the model performance, Shu \etal~\cite{shu2018compressing} proposed to construct the embeddings with few basis vectors. For each word, the composition of basis vectors is determined by a hash code.

On the other hand, Hinton \etal~\cite{hinton2011transforming} presented capsule, which is a small group of neurons. The activities of neurons are used to represent the various properties of an entity. Sabour \etal~\cite{sabour2017dynamic} applied this concept to neural network firstly, a novel routing algorithm called dynamic routing was adopted to select active capsules. The experiments of Capsule Networks (CapsNets) showed capsules could learn a more robust representation than Convolutional Neural Networks (CNNs) in image classification task.

In this paper, we aim to reduce the number of parameters used in word embeddings, while maintaining a competitive accuracy compared to the state-of-the-art approach. To do so, we propose a Compositional Weighted Coding (CWC) embedding to reduce the number of parameters. Then we apply CapsNet to the classification of texts, and propose a novel and robust routing algorithm named k-means routing to determine the connection strength between lower-level and upper-level capsules. 

Our CWC embedding significantly differs from the method proposed by Shu \etal~\cite{shu2018compressing}. The work by Shu \etal~\cite{shu2018compressing} selects exclusive codeword vector in each codebook, while our method uses all the codeword vectors in each codebook and then weights them to form the word embedding. To distinguish with the Compositional Coding Embedding (CC Embedding) method proposed by Shu \etal~\cite{shu2018compressing}, we call our compositional coding method as Compositional Weighted Coding Embedding (CWC Embedding). And our k-means routing uses cosine similarity to obtain the coupling coefficient between lower-level and upper-level capsules, while the dynamic routing proposed by Sabour \etal~\cite{sabour2017dynamic} uses dot product value to determine the coupling coefficient. Furthermore, the coefficient update strategies are also different.

The main contributions of this work are three-folds. First, we propose a new compositional coding approach for constructing the word embeddings with significantly fewer parameters than conventional approach. Second, we propose a novel routing method named k-means routing to decide the connection degree between lower-level and upper-level capsules, it is more stable and robust than dynamic routing. Third, we construct an end-to-end CapsNet with Bidirectional Gated Recurrent Units (BiGRUs,~\cite{schuster1997bidirectional, cho2014properties}) for text classification and achieve comparable results to the state-of-the-art method.

\section{Related Work}

\begin{figure}
  \centering
  \includegraphics[width=\linewidth]{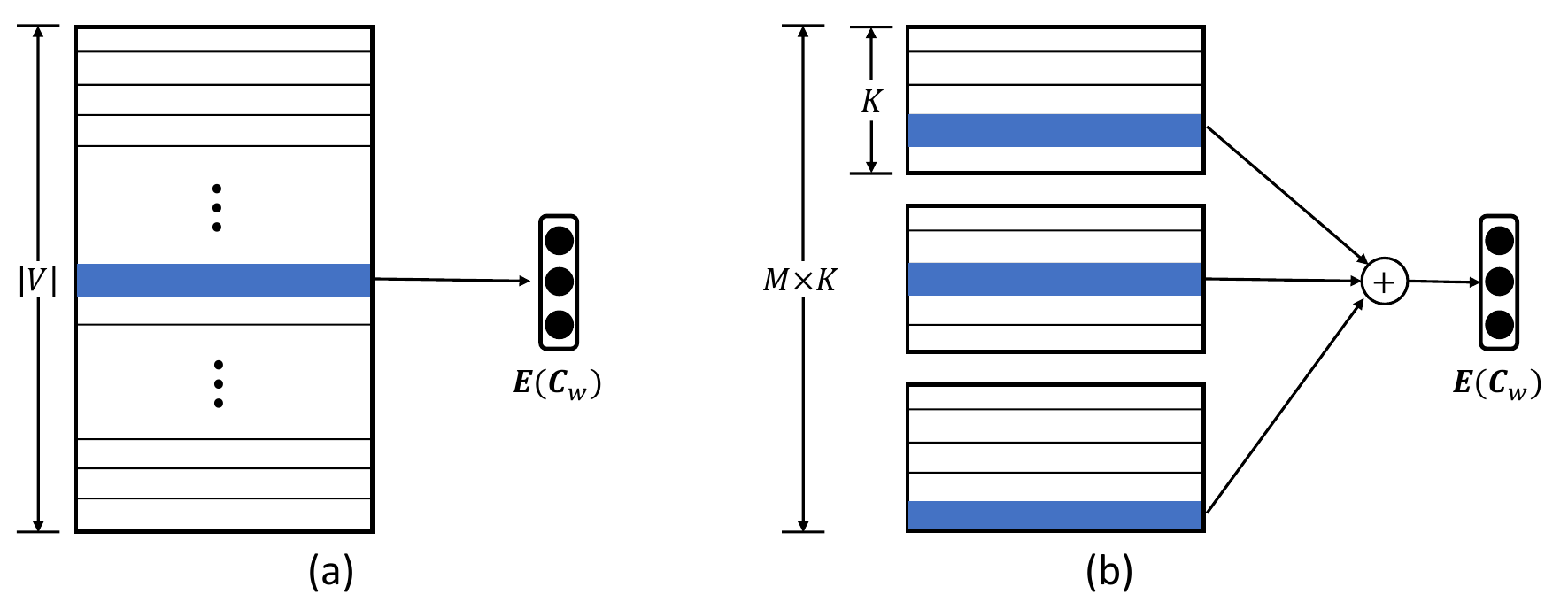}
  \caption{Comparison of embedding computations between the conventional approach (a) and compositional coding approach (b) for constructing embedding vectors.}
  \label{embedding_coding}
\end{figure}

\subsection{Compositional Coding Embedding}

Word embeddings play an important role in NLP models. Neural word embeddings encapsulate the linguistic information of words in continuous vectors. However, as each word is assigned an independent embedding vector, the number of parameters in the embedding layer can be huge. For example, when each embedding has 500 dimensions, the network has to hold 100M embedding parameters to represent 200K words.

Shu \etal~\cite{shu2018compressing} hypothesized that learning independent embeddings causes more redundancy in the embedding vectors, as the inter-similarity among words is ignored. Some words are very similar regarding the semantics. For example, ``dog'' and``dogs'' have almost the same meaning, except one is plural. To efficiently represent these two words, it is desirable to share information between the two embeddings.

Following the intuition of creating partially shared embeddings, instead of assigning each word a unique ID, Shu \etal~\cite{shu2018compressing} represent each word $w$ with a $M$ dimensional code $\mathbf{C}_w=(\mathbf{C}_w^1,\mathbf{C}_w^2,\dots,\mathbf{C}_w^M)$. Each component $\mathbf{C}_w^i$ is an integer number in $[1,K]$. Ideally, similar words should have similar codes. For example, we may desire $\mathbf{C}_{dog}=(3,2,4,1)$ and $\mathbf{C}_{dogs}=(3,2,4,2)$. Once we have obtained such compact codes for all words in the vocabulary, we use embedding vectors to represent the codes rather than the unique words. More specifically, we create $M$ codebooks $\mathbf{E}_1,\mathbf{E}_2,\dots,\mathbf{E}_M$, each containing $K$ codeword vectors. The embedding of a word is computed by summing up the codewords corresponding to all the components in the code as
    \begin{align}
      \mathbf{E}(\mathbf{C}_w) = \sum_{i=1}^M \mathbf{E}_i(\mathbf{C}_w^i)
    \end{align}
where $\mathbf{E}_i(\mathbf{C}_w^i)$ is the $\mathbf{C}_w^i$-th codeword in the codebook $\mathbf{E}_i$. In this way, the number of vectors in the embedding layer will be $M \times K$, which is usually much smaller than the vocabulary size ($\vert V \vert$). Fig.~\ref{embedding_coding} gives an intuitive comparison between the compositional approach and the conventional approach (assigning unique IDs).

Shu \etal~\cite{shu2018compressing} used this method to compress NLP models, and proposed to directly learn the discrete codes in an end-to-end neural network by applying the Gumbel-softmax~\cite{maddison2017concrete, jang2017categorical} trick. Experiments showed the compression rate achieves $98\%$ in a sentiment analysis task and $94\% \sim 99\%$ in machine translation tasks without performance loss.

Inspired by this, we introduce the compositional coding mechanism in our model to generate the word embeddings. And to make sure the model could be trained end-to-end without applying the Gumbel-softmax trick, we propose a variant of Compositional Coding Embedding (CC Embedding) method called Compositional Weighted Coding Embedding (CWC Embedding). The CWC Embedding method requires fewer parameters than conventional approach, and has no restrictions on code. It is more flexible than CC Embedding.

\subsection{Capsule Network}

The concept of capsules is invented by Hinton \etal~\cite{hinton2011transforming} and used recently~\cite{sabour2017dynamic, hinton2018matrix}. CapsNet is designed for image feature extraction, it is developed based on CNN. However, unlike traditional CNN, in which the presence of feature is represented with scalar value in feature maps, the features in CapsNet are represented with capsules (aka vectors). In the work of Sabour \etal~\cite{sabour2017dynamic}, the direction of capsules reflects the properties of the features and the length ($L_2$ norm) of capsules reflects the probability of the presence of different features. The transmission of information between layers follows dynamic routing mechanism. The specific procedure of dynamic routing can be found in~\cite{sabour2017dynamic}.

Yang \etal~\cite{yang2018investigating} explored capsule networks with dynamic routing for text classification, and proposed three strategies to stabilize the dynamic routing process. Experiments on six small text classification benchmarks showed the effectiveness of capsule networks for text classification. However, the work only conducted experiments on small datasets, and compared with basic CNN-based or LSTM-based methods, without considering the number of parameters. To make up for the deficiency, we conducted experiments on eight challenging text classification benchmarks with our model, and combined with the proposed CWC embedding to save parameters.

Inspired by CapsNet, the capsule mechanism is adopted in our model to generate class capsules on the basis of feature capsules. The feature capsules are extracted from BiGRUs. A variant of dynamic routing called k-means routing is proposed to update weights between capsules from one layer to the next layer so that the properties captured by feature capsules can be propagated to suitable class capsules. Thus, each text is modeled as multiple feature capsules, and then modeled as multiple class capsules. Different feature capsules reflect the properties of the feature from different aspects.

\section{Proposed Methods}

In this section, we describe the compositional weighted coding embedding approach, k-means routing algorithm and the end-to-end model in details.

\begin{figure}
  \centering
  \includegraphics[width=\linewidth]{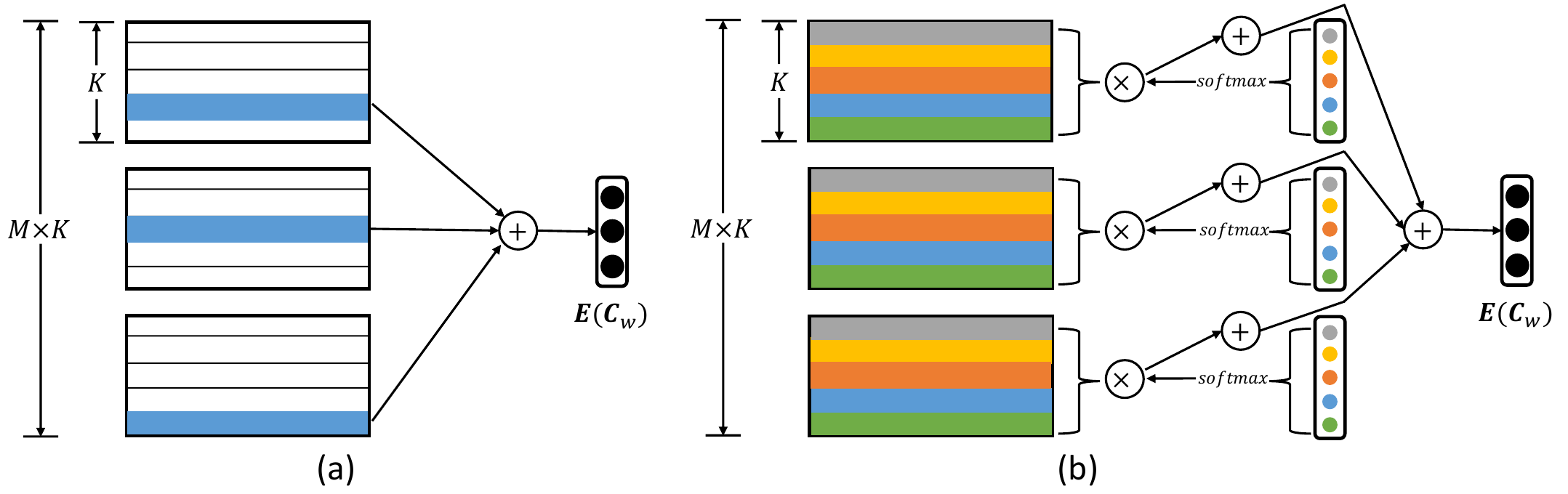}
  \caption{Comparison of embedding computations between the CC Embedding (a) and CWC Embedding (b) for constructing embedding vectors.}
  \label{embedding_layer}
\end{figure}

\subsection{Compositional Weighted Coding Embedding}

Unlike the CC Embedding method, which selects exclusive codeword vector in each codebook, CWC Embedding method uses all codeword vectors in each codebook to form the word embedding. And CC Embedding method restricts that the code must be integer number, our CWC Embedding method eliminates this limitation.

Suppose the vocabulary size is $\vert V \vert$, we create $M$ codebooks $\mathbf{E}_1, \mathbf{E}_2,\dots,\mathbf{E}_M$, each containing $K$ codeword vectors. For CWC Embedding, the embedding of word $w$ is computed by summing up the weighted codewords corresponding to all the components in the code as
    \begin{align}
    \label{formula:cwc}
      \mathbf{E}(\mathbf{C}_w) = \sum_{i=1}^M \sum_{j=1}^K \mathop{softmax}\limits_j (\mathbf{C}_w^{ij})\mathbf{E}_i(j)
    \end{align}
where $\mathbf{E}_i(j)$ is the $j$-th codeword in the codebook $\mathbf{E}_i$, $\mathbf{C}_w^{ij}$ is the $j$-th code for the codebook $\mathbf{E}_i$. From the Formula (\ref{formula:cwc}), we can see the code does not need to be integer number. Fig.~\ref{embedding_layer} gives an intuitive comparison between the CC Embedding and the CWC Embedding. 

Moreover, $M$ and $K$ are hyper-parameters, they are designated by user in the CC Embedding. However, in the CWC Embedding, only $M$ need to be designated, $K$ is determined as follows
    \begin{align}
      K = \lceil\sqrt[M]{\vert V \vert}\rceil
    \end{align}
because $K^M$ is the total number of all the combination of codeword vectors, it makes sure $K^M \geq \vert V \vert$, which means each word can be assigned with a unique combination of codeword vectors.

From the process of word embedding computation, we can see our CWC Embedding method requires fewer parameters than conventional method, just as the CC Embedding method. The more detailed results about parameters are presented in Section~\ref{sec:ablation}.

\subsection{K-means Routing}
\label{sec:k_means}

The capsule layer receives lower-level capsules, which represent low-level features, then the routing algorithm clusters the low-level features to high-level features. We know that k-means clustering is an efficient method to cluster features, and produce a cluster centroid by using all the clustered features. Based on this, we propose k-means routing. We regard the k-means routing algorithm between $l^{th}$ layer's capsules and $(l+1)^{th}$ layer's capsules as a k-means clustering process. The $(l+1)^{th}$ layer's capsules are regarded as the cluster centers of the $l^{th}$ layer's capsules.

Then we briefly review k-means clustering and its optimization procedure. Given $n$ capsules $\mathbf{u}_1,\dots,\mathbf{u}_n$ and the metric $d$, k-means clustering is to find $k$ cluster centers $\mathbf{v}_1,\dots,\mathbf{v}_k$ to minimize the following loss function:
    \begin{align}
      L=\sum_{i=1}^n \min_{j=1}^k d(\mathbf{u}_i, \mathbf{v}_j)
    \end{align}
so the objective function is
    \begin{align}
      \mathbf{v}_1,\dots,\mathbf{v}_k = \mathop{\arg\min}_{\mathbf{v}_1,\dots,\mathbf{v}_k}L = \mathop{\arg\min}_{\mathbf{v}_1,\dots,\mathbf{v}_k}\sum_{i=1}^n \min_{j=1}^k d(\mathbf{u}_i, \mathbf{v}_j)
    \end{align}
Because the objective function contains $\min$ operation, it is not smooth and does not always have a gradient. Then we use the approximation~\cite{cook2011basic}
    \begin{align}
      \max(\lambda_1,\dots,\lambda_n) & = \lim_{K\to+\infty}\frac{1}{K}\ln\left(\sum_{i=1}^n e^{\lambda_i K}\right) \notag \\ & \approx \frac{1}{K}\ln\left(\sum_{i=1}^n e^{\lambda_i K}\right)
    \end{align}
and
    \begin{align}
      \min(\lambda_1,\dots,\lambda_n) = -\max(-\lambda_1,\dots,-\lambda_n)
    \end{align}
to smooth the loss function
    \begin{align}
      L\approx -\frac{1}{K}\sum_{i=1}^n \ln\left(\sum_{j=1}^k e^{-K\cdot d(\mathbf{u}_i, \mathbf{v}_j)}\right) = -\frac{1}{K}\sum_{i=1}^n\ln Z_i
    \end{align}
It is a differentiable function, so we can get its gradient
    \begin{align}
      \frac{\partial L}{\partial \mathbf{v}_j}\approx \sum_{i=1}^n \frac{e^{-K\cdot d(\mathbf{u}_i, \mathbf{v}_j)}}{Z_i} \frac{\partial d(\mathbf{u}_i, \mathbf{v}_j)}{\partial \mathbf{v}_j} = \sum_{i=1}^n c_{ij}\frac{\partial d(\mathbf{u}_i, \mathbf{v}_j)}{\partial \mathbf{v}_j}
    \end{align}
where
	\begin{align}
     c_{ij} = \mathop{softmax}_j\Big(-K\cdot d(\mathbf{u}_i, \mathbf{v}_j)\Big)
    \end{align}

For obtaining $\mathbf{v}_j$, we need to solve the equations $\partial L/\partial \mathbf{v}_j=0$, which is non-linear mostly and can not be solved analytically. So we need introduce an iterative process, suppose $\mathbf{v}_j^{(r)}$ is the result of $\mathbf{v}_j$ after $r$ iterations, then we could let
    \begin{align}
      \sum_{i=1}^n c_{ij}^{(r)}\frac{\partial d\left(\mathbf{u}_i, \mathbf{v}_j^{(r+1)}\right)}{\partial \mathbf{v}_j^{(r+1)}}=0
    \end{align}
for the routing between capsules, we use the following metric:
    \begin{align}
      d(\mathbf{u}_i, \mathbf{v}_j)=-\left\langle\frac{\mathbf{u}_i}{\Vert\mathbf{u}_i\Vert}, \frac{\mathbf{v}_j}{\Vert\mathbf{v}_j\Vert}\right\rangle
    \end{align}
here $\langle \cdot, \cdot \rangle$ is the scalar product operation. We can simply take
    \begin{align}
      \mathbf{v}_j^{(r+1)}=\sum\limits_{i=1}^n c_{ij}^{(r)}\mathbf{u}_i   
    \end{align}
here $c_{ij}^{(r)} = \mathop{softmax}\limits_j \Big(\langle\frac{\mathbf{u}_i}{\Vert\mathbf{u}_i\Vert}, \frac{\mathbf{v}_j^{(r)}}{\Vert\mathbf{v}_j^{(r)}\Vert}\rangle\Big)$, it means $\mathbf{v}_j^{(r+1)}$ is the sum of those nearest $\mathbf{u}$s to $\mathbf{v}_j^{(r)}$. 

Finally, to achieve a complete routing algorithm, we need to solve these three problems: how to initialize the cluster centers, how to identify capsules at different position, how to guarantee the cluster centers keep the main information of input features. They all can be solved by inserting transformation matrix $\mathbf{W}_{ij}$:
    \begin{align}
      \mathbf{v}_j^{(r+1)}= \sum\limits_{i=1}^n c_{ij}^{(r)}\mathbf{W}_{ij}\mathbf{u}_i
    \end{align}
here $c_{ij}^{(r)}=\mathop{softmax}\limits_j \Big(\langle\frac{\mathbf{W}_{ij}\mathbf{u}_i}{\Vert\mathbf{W}_{ij}\mathbf{u}_i\Vert}, \frac{\mathbf{v}_j^{(r)}}{\Vert\mathbf{v}_j^{(r)}\Vert}\rangle\Big)$. For the simplicity of this iterative process, we assign the sum of $\mathbf{u}_i$ averagely to each cluster center as $\mathbf{v}_j^{(0)}$. Because we want to use the length of capsule to represent the probability that a category's entity exists, a $squash$ function has been introduced:
    \begin{align}
      squash(\mathbf{v}_j)= \frac{\Vert \mathbf{v}_j \Vert}{1 + \Vert \mathbf{v}_j \Vert^2} \mathbf{v}_j
    \end{align}
    
The whole procedure is summarized on Algorithm~\ref{alg:A}. Inserting $\mathbf{W}_{ij}$ is a beautiful trick, which induces different cluster centers by one same initialization method. In addition, $\mathbf{W}_{ij}$ can keep the position information and increase or decrease dimension of capsule, which means the cluster centers have enough representation ability.

\begin{algorithm}
  \caption{K-means Routing}
  \label{alg:A}
  \begin{algorithmic}[1]
    \Procedure {Routing}{$\mathbf{u}_i, r$}
    \State Initialize $\mathbf{v}_j \leftarrow \frac{1}{k} \sum\limits_{i=1}^n \mathbf{W}_{ij} \mathbf{u}_i$
	\For {$r$ iterations}
	 \State $b_{ij} \leftarrow \langle\frac{\mathbf{W}_{ij}\mathbf{u}_i}{\Vert\mathbf{W}_{ij}\mathbf{u}_i\Vert}, \frac{\mathbf{v}_j}{\Vert\mathbf{v}_j\Vert}\rangle$
      \State $c_{ij} \leftarrow \mathop{softmax}\limits_j b_{ij}$
	  \State $\mathbf{v}_j \leftarrow \sum\limits_{i=1}^n c_{ij} \mathbf{W}_{ij} \mathbf{u}_i$
	\EndFor
	\State \Return $squash(\mathbf{v}_j)$
	\EndProcedure
  \end{algorithmic}
\end{algorithm}

K-means routing is similar to dynamic routing in general, but it has three differences. First of all, we do not apply the $squash$ function to capsule $\mathbf{v}_j$ in the period of iteration, we just squash it after iteration. Secondly, $b_{ij}$ is replaced by new $b_{ij}$, however, in dynamic routing, $b_{ij}$ is replaced by new $b_{ij}$ plus old $b_{ij}$. This is the biggest difference between our routing algorithm and dynamic routing. Finally, the cosine similarity is computed between $\mathbf{v}_j$ and $\mathbf{W}_{ij} \mathbf{u}_i$ instead of dot product.

According to the $b_{ij}$ update step as described in dynamic routing, after $r$ iterations:
    \begin{align}
      {\mathbf{v}_j}^{(r)} \sim squash\left(\sum_{i} \frac{e^{r \hat{\mathbf{u}}_i \cdot \mathbf{v}_j}}{Z_i} \hat{\mathbf{u}}_i \right)
    \end{align}
where 
    \begin{align}
      Z_i = \sum_{j} e^{\hat{\mathbf{u}}_i \cdot \mathbf{v}_j}, \quad \hat{\mathbf{u}}_i = \mathbf{W}_{ij} \mathbf{u}_i
    \end{align}
if $r \rightarrow + \infty $, we find the result of $\mathop{softmax}$ will be either $0$ or $1$. In other words, each lower capsule is linked to sole upper capsule. This is unreasonable, we know there are common characteristics among different categories, so we hope the common characteristics can be linked to all those categories. This is why we do not plus old $b_{ij}$ when we update $b_{ij}$. Section~\ref{sec:ablation} shows the comparison results between k-means routing and dynamic routing.

\begin{figure}
  \centering
  \includegraphics[width=\linewidth]{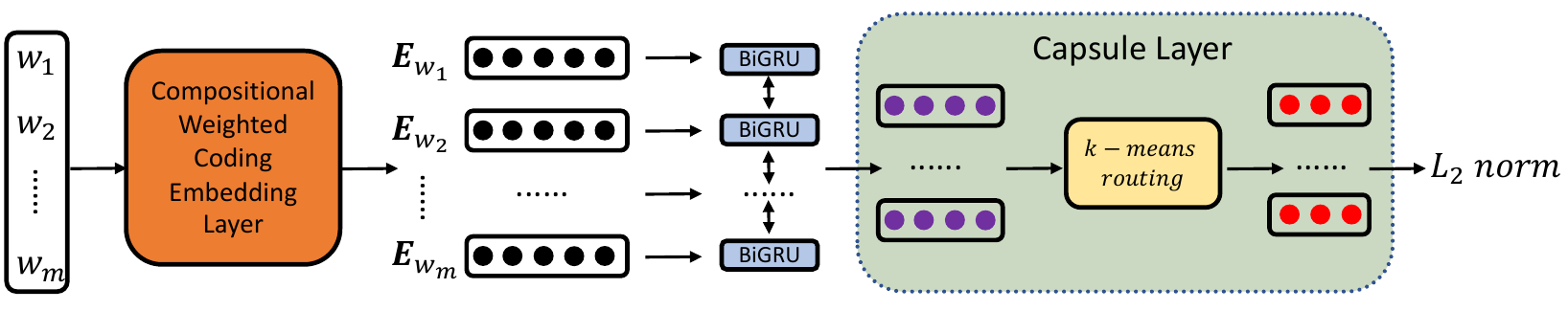}
  \caption{Our model structure. We use $L_2$ norm of capsule to represent the probability that a category's entity exists.}
  \label{model_structure}
\end{figure}

\subsection{Model Architecture}

Our architecture consists of a CWC Embedding layer, a BiGRU layer and a capsule layer. The aim of CWC Embedding layer is to obtain the word embedding. BiGRU layer extracts lower features and feeds them into capsule layer, which takes advantage of them to form the upper features and correctly classify the texts.

We design a simple model to validate the effectiveness and test the performance of our approach. The structure is illustrated in Fig.~\ref{model_structure}. The number of codebook is 8, and the embedding dimension is 64. The hidden size of BiGRU is 128, and the number of recurrent layers is 2. We introduce Dropout~\cite{hinton2012improving} on the outputs of each RNN layer except the last layer, with dropout probability equal to 0.5. The dimension of lower capsules is 8, and the dimension of upper capsules is 16. As the number of categories is related to specific dataset, so the number of upper capsules is specified with particular dataset.

\begin{table*}
  \centering
  \caption{Test accuracy (\%) on eight text classification datasets. The methods with * indicate using fine-tuning, - means the results are not reported on their papers. The best results are bold.}
  \label{table:sota}
  \begin{tabular}{l|cccccccc}
    \toprule
    Method  & AG  & DBP  & Yah.A  & Sogou  & Yelp.P  & Yelp.F & Amz.P  & Amz.F \\
    \midrule
    DPCNN*~\cite{johnson2017deep}    & 93.13  & 99.12  & 76.10 & 96.52  & 97.36  & 69.42 & 96.68 & 65.19 \\
    ULMFiT*~\cite{howard2018universal}    & 94.99  & 99.20  & - & -  & 97.84  & 70.02 & - & - \\
    Region Emb.~\cite{qiao2018new}    & 92.80  & 98.90  & 73.70 & \textbf{97.60}  & 96.40  & 64.90 & 95.10 & 60.90 \\
    BERT-FiT*~\cite{sun2019fine}    & 94.75  & 99.29  & \textbf{77.58} & 97.57  & 97.72  & 69.94 & - & - \\
    BERT-Large*~\cite{xie2020unsupervised}    & -  & 99.36  & - & -  & 98.11  & 70.68 & 97.37 & 65.83 \\
    XLNet*~\cite{yang2019xlnet}    & \textbf{95.55}  & \textbf{99.40}  & - & -  & \textbf{98.63}  & \textbf{72.95} & \textbf{97.89} & \textbf{68.33} \\
    \midrule
    Ours    & 92.39         & 98.72       & 73.85      & 97.25   & 96.48  & 65.85 & 94.96 & 60.95 \\
    \bottomrule
  \end{tabular}
\end{table*}

\begin{table*}
  \centering
  \caption{The number of model parameters (M) on eight text classification datasets. The minimum number of model parameters are bold.}
  \label{table:sota_parameter}
  \begin{tabular}{l|cccccccc}
    \toprule
    Method  & AG  & DBP  & Yah.A  & Sogou  & Yelp.P  & Yelp.F & Amz.P  & Amz.F \\
    \midrule
    DPCNN~\cite{johnson2017deep}    & 4.24  & 35.34  & 49.64 & 7.05  & 13.09  & 14.13 & 59.84 & 53.73 \\
    ULMFiT~\cite{howard2018universal}    & 20.56  & 51.65  & 65.96 & 23.37  & 29.41  & 30.45 & 76.16 & 70.05 \\
    Region Emb.~\cite{qiao2018new}    & 43.81  & 233.33  & 370.61 & 101.78  & 118.07  & 127.26 & 403.85 & 364.86 \\
    BERT-FiT~\cite{sun2019fine}    & 78.59  & 451.69  & 623.32 & 112.27  & 184.77  & 197.21 & 745.78 & 672.47 \\
    BERT-Large~\cite{xie2020unsupervised}    & 169.65 & 667.11  & 895.95 & 214.55  & 311.22  & 327.80 & 1,059.23 & 961.49 \\
    XLNet~\cite{yang2019xlnet}    & 193.26  & 690.72  & 919.57 & 238.16  & 334.83  & 351.42 & 1,082.84 & 985.10 \\
    \midrule
    Ours    & \textbf{2.46}         & \textbf{26.80}       & \textbf{37.52}      & \textbf{4.71}   & \textbf{8.48}  & \textbf{9.14} & \textbf{45.15} & \textbf{40.58} \\
    \bottomrule
  \end{tabular}
\end{table*}

\begin{table}
  \centering
  \caption{Statistics of the benchmark text classification datasets.}
  \label{table:datasets}
  \begin{tabular}{c|rrrr}
    \toprule
    Dataset   & \# Class  & \# Train & \# Test & $\vert V \vert$ \\
    \midrule
    AG    & 4        & 120k    & 7.6k   & 62,535\\
    DBP   & 14       & 560k    & 70k    & 548,338 \\
    Yah.A  & 10       & 1,400k  & 60k    & 771,820  \\ 
    Sogou   & 5        & 450k    & 60k    & 106,385  \\
    Yelp.P    & 2        & 560k    & 38k    & 200,790  \\
    Yelp.F    & 5        & 650k    & 50k    & 216,985 \\ 
    Amz.P  & 2        & 3,600k  & 400k   & 931,271 \\
    Amz.F  & 5        & 3,000k  & 650k   & 835,818  \\
    \bottomrule
  \end{tabular}
\end{table}

\section{Experiments}

In this section, we conduct extensive experiments on the 8 benchmark text classification datasets. We first introduce the details of the 8 datasets, and present the experimental settings. Then we conduct experiments with our model and compare with the state-of-the-art methods.

\subsection{Datasets}

We use publicly available datasets from Zhang \etal~\cite{zhang2015character} to evaluate our model. These datasets are AG's News (AG), DBPedia (DBP), Yahoo! Answers (Yah.A), Sogou News (Sogou), Yelp Review Polarity (Yelp.P), Yelp Review Full (Yelp.F), Amazon Review Polarity (Amz.P) and Amazon Review Full (Amz.F). Table~\ref{table:datasets} shows summary of these datasets' main features.

The datasets used in this paper not only contain English words, but also contain digital numbers, punctuations, Chinese words, etc. So we need to preprocess them to make sure the preprocessed samples can be tokenized by space character easily. To achieve this goal, we turn the sentences to be lower case firstly, then add one space character before and after the words or characters which are not belonging to English words. Additionally, we just take the first 5,000 words into our model.

\subsection{Experimental Settings}

We implement our model with PyTorch library~\cite{paszke2019pytorch}, all the experiments are performed on a single NVIDIA Tesla V100 GPU. The number of routing iterations is fixed to 3, and the batch size is 32. We train the model with 10 epochs. A lot of loss functions have been experimented, finally we decided using a compositional loss function of margin loss~\cite{sabour2017dynamic} and focal loss~\cite{lin2017focal} to compute our model's loss:
    \begin{align}
      L = \mathcal{L}_{m} + \mathcal{L}_{f}
    \end{align}
    where margin loss is
    \begin{align}
      \mathcal{L}_{m} & = \frac{1}{k} \sum_{j=1}^k \Big( T_j \max(0, 0.9-\Vert \mathbf{v}_j \Vert)^2 \notag \\ & + 0.5(1-T_j) \max(0, \Vert \mathbf{v}_j \Vert-0.1)^2 \Big)    	
    \end{align}
    and focal loss is
    \begin{align}
      \mathcal{L}_{f} = - 0.25(1-\Vert \mathbf{v}_j \Vert)^2 \log \Vert \mathbf{v}_j \Vert \Big)    	
    \end{align}
where $T_j = 1$ iff a text of class $j$ is present, the hyper-parameters follow the original paper. It is optimized through ADAM scheme~\cite{kingma2015adam} with learning rate $0.001$. The ablation study of loss function can be found in Section~\ref{sec:ablation}. The code of our work is available on~\url{https://github.com/leftthomas/CCCapsNet}.

\subsection{Comparison with SOTA}

We compare the proposed method against existing state-of-the-art methods on the eight text classification datasets. The quantitative results of different methods are reported in Table~\ref{table:sota}. Please note our model is trained from scratch. From this table we can see that our method achieves competitive accuracy compared with Region Emb.~\cite{qiao2018new}, which is also trained from scratch. And compared with the methods which are pre-trained on other text datasets, and then fine-tuned on the target dataset, such as BERT-FiT~\cite{sun2019fine}, the performance gap of our method is not large.

We also provide the number of model parameters of theses methods in Table~\ref{table:sota_parameter}. Combined with Table~\ref{table:sota}, we can easily conclude that our method greatly reduces the overall parameters of the model without much loss of performance. For example, our model only uses 26.80M parameters, but ULMFiT~\cite{howard2018universal} holds 51.65M parameters on DBPedia dataset, which is around two times as ours. And the performance gap between this two methods is only 0.48\%. For XLNet~\cite{yang2019xlnet}, the number of parameters of our model is only 1.27\% of its on AG's News dataset, and the compression ratio is almost 100 times. Looking at all datasets and comparing with other methods, it is not difficult to see that our method has made great compression in the amount of model parameters, but the performance impact is minimal.

\begin{table}
  \centering
  \caption{Ablation studies for the three loss functions: cross entropy loss ($\mathcal{L}_{c}$), focal loss ($\mathcal{L}_{f}$) and margin loss ($\mathcal{L}_{m}$) on AG's News dataset. The best test accuracy (\%) are bold.}
  \label{table:loss}
  \scalebox{0.79}{
  \begin{tabular}{ccccccc}
    \toprule
    $\mathcal{L}_{c}$  & $\mathcal{L}_{f}$   & $\mathcal{L}_{m}$  & $\mathcal{L}_{c}$ + $\mathcal{L}_{f}$  & $\mathcal{L}_{c}$ + $\mathcal{L}_{m}$   & $\mathcal{L}_{f}$ + $\mathcal{L}_{m}$ & $\mathcal{L}_{c}$ + $\mathcal{L}_{f}$ + $\mathcal{L}_{m}$   \\ \midrule
    92.05 & 92.13  & 92.37   & 92.09 & 91.95 & \textbf{92.64} &  92.38  \\
    \bottomrule
  \end{tabular}
  }
\end{table}

\subsection{Ablation Studies}
\label{sec:ablation}

\textbf{Selection of loss function}. In order to make our method comparable to the state-of-the-art methods and determine which loss function is suitable for our method, we also explore three mainstream loss functions and their combinations, and conduct relevant comparative experiments on AG's News dataset. The experimental results are shown in Table~\ref{table:loss}, the loss functions adopted are cross entropy loss, focal loss and margin loss.

From the results, we can see that focal loss and margin loss are better than cross entropy loss, and combining these two loss functions can achieve the best accuracy. We can also find that in some cases, the combination of cross entropy loss function will even bring negative effects, making the accuracy lower, \eg, $\mathcal{L}_{m}$ \textit{vs} $\mathcal{L}_{c}$ + $\mathcal{L}_{m}$. Based on this experiment, we finally determine to use margin loss + focal loss as our model's loss function.

\begin{table}
  \centering
  \caption{Ablation studies for routing algorithms on eight text classification datasets. The best test accuracy (\%) are bold.}
  \label{table:routing}
  \scalebox{0.71}{
  \begin{tabular}{l|cccccccc}
    \toprule
    Algorithm  & AG  & DBP  & Yah.A  & Sogou  & Yelp.P  & Yelp.F & Amz.P  & Amz.F \\
    \midrule
    Dynamic    & 92.14  & 98.70  &  \textbf{73.98} & 97.23  & 96.34  & 65.81 & 94.87 & 60.90 \\
    K-means    & \textbf{92.39}  & \textbf{98.72}  & 73.85 & \textbf{97.25}  & \textbf{96.48}  & \textbf{65.85} & \textbf{94.96} & \textbf{60.95} \\
    \bottomrule
  \end{tabular}
  }
\end{table}

\noindent\textbf{Effectiveness of k-means routing}. Section~\ref{sec:k_means} shows that k-means routing is more stable and robust than dynamic routing in theory, here we give experiments to verify this. The quantitative results are shown in Table~\ref{table:routing}. 

In order to prove the generality of the conclusion, we have conducted experiments on eight text datasets. From the results, we can see that our method has achieved the best accuracy on seven of them, except Yahoo! Answers dataset. From this, we can conclude that our routing algorithm is significantly better than the dynamic routing algorithm. This conclusion derived from the experiments further complements our theoretical analysis in Section~\ref{sec:k_means}.

\begin{table*}
  \centering
  \caption{Ablation studies for the components of our proposal on eight text classification datasets. The model contains a conventional embedding layer, a BiGRU layer and a fully connected layer is used as our baseline. The best test accuracy (\%) are bold.}
  \label{table:model}
  \begin{tabular}{c|ccc|cccccccc}
    \toprule
    Exp & CC  & CWC   & Capsule  & AG  & DBP   &  Yah.A   & Sogou  & Yelp.P & Yelp.F & Amz.P & Amz.F \\ \midrule
    1 & - & -           & -          & \textbf{92.64}         & 98.84       & \textbf{74.13}      & 97.37   & \textbf{96.69}  & \textbf{66.23} & 95.09 & 60.78  \\
    2 & \checkmark & -  & -          & 83.13         & 96.06       & 57.87      & 95.20   & 92.37  & 56.66 & 89.04 & 51.30    \\
    3 & - & \checkmark  & - & 91.93         & 98.83       & 73.58      & 97.37   & 96.35  & 65.11 & 94.90 & 60.29 \\
    4 & - & -  & \checkmark          & 92.18         & \textbf{98.86}       & 74.12      & \textbf{97.52}   & 96.56  & \textbf{66.23} & \textbf{95.18} & \textbf{61.36} \\
    5 & \checkmark & -  & \checkmark & 84.05         & 96.27       & 60.31      & 96.00   & 92.82  & 59.48 & 89.07 & 52.06  \\
    6 & - & \checkmark  & \checkmark & 92.39         & 98.72       & 73.85      & 97.25   & 96.48  & 65.85 & 94.96 & 60.95  \\ 
    \bottomrule
  \end{tabular}
\end{table*}

\noindent\textbf{Effectiveness of CWC Embedding}. To find out how much each component of our proposed method contributes to the overall performance, we conduct a series of experiments, and the relevant quantitative results are shown in Table~\ref{table:model}. At the same time, we also provide the corresponding model parameters, as shown in Table~\ref{table:model_parameter}. To make a fair comparison, we design a simple model as our baseline, which has a conventional embedding layer, a BiGRU layer and a fully connected layer. On this basis, we replace the conventional embedding layer and the fully connected layer to explore the role of each component. At the same time, we also give the results of various combinations to help us better understand the role of each component. Here, we focus on the role of CWC Embedding, the role of capsule will be explained in the next part.

Look at the first three rows of Table~\ref{table:model}, the first row is the result of baseline model, the second row is the result of the model which replaces the conventional embedding layer with CC Embedding, and the third row is the result of the model which replaces the conventional embedding layer with our CWC Embedding. From the results we can see the performance of our CWC Embedding is very close to the baseline model. And look at the first three rows of Table~\ref{table:model_parameter}, we can observe that our CWC Embedding requires fewer parameters than the baseline model on all the eight datasets. For example, the number of model parameters on AG's News dataset of our CWC Embedding is 2.45M, but the baseline model needs 4.45M parameters, which is around 1.8 times than ours. For the other datasets, our CWC Embedding can save about $24.68\% \sim 36.29\% $ parameters. 

As for CC Embedding, although it saves parameters like our CWC Embedding, its performance drops sharply (\ie, $74.13\% \rightarrow 57.87\%$ on Yahoo! Answers dataset). Then we can conclude that our CWC Embedding is more suitable compared with CC Embedding to replace the conventional embedding layer. This also verifies our proposition that our CWC Embedding can maintain competitive accuracy with significantly fewer parameters.

\begin{table*}
  \centering
  \caption{Ablation studies for the components of our proposal on eight text classification datasets. The model contains a conventional embedding layer, a BiGRU layer and a fully connected layer is used as our baseline. The minimum number of model parameters (M) are bold.}
  \label{table:model_parameter}
  \begin{tabular}{c|ccc|cccccccc}
    \toprule
    Exp & CC  & CWC   & Capsule  & AG  & DBP   &  Yah.A   & Sogou  & Yelp.P & Yelp.F & Amz.P & Amz.F \\ \midrule
    1 & - & -           & -          & 4.45         & 35.54       & 49.84      & 7.25   & 13.30  & 14.33 & 60.05 & 53.94 \\
    2 & \checkmark & -  & -          & \textbf{2.45}         & \textbf{26.77}       & \textbf{37.50}      & \textbf{4.70}   & \textbf{8.48}  & \textbf{9.13} & \textbf{45.15} & \textbf{40.57}    \\
    3 & - & \checkmark  & - & \textbf{2.45}         & \textbf{26.77}       & \textbf{37.50}      & \textbf{4.70}   & \textbf{8.48}  & \textbf{9.13} & \textbf{45.15} & \textbf{40.57} \\
    4 & - & -  & \checkmark          & 4.46         & 35.57       & 49.86      & 7.26   & 13.30  & 14.34 & 60.05 & 53.95 \\
    5 & \checkmark & -  & \checkmark & 2.46         & 26.80       & 37.52      & 4.71   & \textbf{8.48}  & 9.14 & \textbf{45.15} & 40.58  \\
    6 & - & \checkmark  & \checkmark & 2.46         & 26.80       & 37.52      & 4.71   & \textbf{8.48}  & 9.14 & \textbf{45.15} & 40.58  \\ 
    \bottomrule
  \end{tabular}
\end{table*}

\noindent\textbf{Effectiveness of capsule layer}. Look at the first row and fourth row of Table~\ref{table:model} and Table~\ref{table:model_parameter}, we can find that the model with capsule layer is slightly better than the model with fully connected layer, the number of parameters added is very small and can be ignored.

In the case of combination, specifically, the last two rows in Table~\ref{table:model}, introducing capsule has brought further improvement to the performance of CC Embedding (second row \textit{vs} fifth row) and CWC Embedding (third row \textit{vs} sixth row). Especially for CC Embedding on Yelp Review Full dataset, the performance improvement reaches 2.82\% ($56.66\% \rightarrow 59.48\%$). These experimental results also show that introducing capsule layer to replace the fully connected layer is beneficial to the final performance of our model.

\section{Conclusions}

In this paper, we proposed CWC Embedding, which can maintain competitive accuracy compared with conventional embedding layer with significantly fewer parameters. Then we developed a more robust and stable k-means routing algorithm than dynamic routing, analyzed the limitation of dynamic routing. Finally, introducing the capsule layer further improves the performance of our method. Extensive experiments on text classification task demonstrated the effectiveness of our method, which achieved competitive performance compared to the state-of-the-art methods with significantly fewer parameters.

\bibliographystyle{model2-names}
\bibliography{refs}

\end{document}